\pdfoutput=1
\documentclass{ifacconf}

\usepackage{natbib}        % required for bibliography
\usepackage[dvipsnames]{xcolor}
\usepackage{graphicx}
\usepackage{graphics} % for pdf, bitmapped graphics files
\usepackage[]{units}
\usepackage{soul}
\usepackage{tikz}
\usepackage{pgfplots}
\usetikzlibrary{positioning}
\usetikzlibrary{shapes.geometric}
\usetikzlibrary{shapes.misc}
\usetikzlibrary{shapes,arrows,patterns}
\usetikzlibrary{plotmarks}
\pgfplotsset{compat=1.14}

\usepackage{paralist}
\usepackage{mathtools}

\usepackage{siunitx}
\usepackage{multirow}
\usepackage{epstopdf}
\graphicspath{ {Pictures/} }
\usepackage{multicol}
\usepackage{epstopdf}

\usepgfplotslibrary{patchplots}
\usepackage{epsfig}
\usepackage{float}
\usepackage{amsmath} % assumes amsmath package installed
\usepackage{amssymb}  % assumes amsmath package installed
\usepackage{mathtools}
\graphicspath{ {Pictures/} }
\DeclareGraphicsExtensions{.pdf,.eps}

\usepackage[acronym,nomain,nonumberlist]{glossaries}
\newacronym{ahe}{AHE}{Adaptive Histogram Equalization}
\newacronym{ar}{AR}{Augmented Reality}
\newacronym{clahe}{CLAHE}{Contrast Limited Adaptive Histogram Equalization}
\newacronym{cnn}{CNN}{Convolutional Neural Network}
\newacronym{dl}{DL}{Deep Learning}
\newacronym{dof}{DoF}{Degree of Freedom}
\newacronym{ekf}{EKF}{Extended Kalman Filter}
\newacronym{fps}{fps}{Frame Per Second}
\newacronym{fov}{FoV}{Field of View}
\newacronym{fbe}{FBE}{Forward Backward Envelope}
\newacronym{gnss}{GNSS}{Global Navigation Satellite System}
\newacronym{gps}{GPS}{Global Positioning System}
\newacronym{kf}{KF}{Extended Kalman Filter}
\newacronym{iid}{IID}{Independent Identically Distributed}
\newacronym{imu}{IMU}{Inertial Measurement Unit}
\newacronym{mae}{MAE}{mean absolute error}
\newacronym{mav}{MAV}{Micro Aerial Vehicle}
\newacronym{mhe}{MHE}{Moving Horizon Estimation}
\newacronym{mimo}{MIMO}{Multiple Input Multiple Output}
\newacronym{mlp}{MLP}{MultiLayer Perceptron}
\newacronym{mpc}{MPC}{Model Predictive Control}
\newacronym{msf}{MSF}{Multi Sensor Fusion}
\newacronym{nmhe}{NMHE}{Nonlinear Moving Horizon Estimation}
\newacronym{nmpc}{NMPC}{Nonlinear Model Predictive Control}
\newacronym{nn}{NN}{Nueral Network}
\newacronym{nwu}{NWU}{North-West-Up}
\newacronym{panoc}{PANOC}{Proximal Averaged Newton-type method for Optimal Control}
\newacronym{pdf}{PDF}{Probability Density Function}
\newacronym{pid}{PID}{Proportional Integral Derivative}
\newacronym{relu}{ReLU}{Rectified Linear Unit}
\newacronym{rmse}{RMSE}{Root Mean Square Error}
\newacronym{rl}{RL}{Reinforcement Learning}
\newacronym{ros}{ROS}{Robot Operating System}
\newacronym{sfm}{SfM}{Structure from Motion}
\newacronym{sqp}{SQP}{Sequential Quadratic Programming}
\newacronym{ugv}{UGV}{Unmanned Ground Vehicle}
\newacronym{ukf}{UKF}{Unscented Kalman Filter}
\newacronym{vi}{VI}{Visual Inertia}
\newacronym{vo}{VO}{Visual Odometry}

\usepackage{bm}

\newlength\fwidth

\begin{document}
\begin{frontmatter}

\title{Subterranean MAV Navigation based on Nonlinear MPC with Collision Avoidance Constraints\thanksref{footnoteinfo}} 

\thanks[footnoteinfo]{This work has been partially funded by the European Unions Horizon 2020 Research and Innovation Programme under the Grant Agreement No. 730302 SIMS. Corresponding author's e-mail: sinsha@ltu.se}

\author[First]{Sina Sharif Mansouri} 
\author[First]{Christoforos Kanellakis} 
\author[Second]{Emil Fresk} 
\author[First]{Bj\"orn Lindqvist} 
\author[First]{Dariusz Kominiak} 
\author[First]{Anton Koval} 
%\author[Third]{Joel Burdick} 
\author[Forth]{Pantelis Sopasakis} 
\author[First]{George Nikolakopoulos} 

\address[First]{Robotics Team, Department of Computer, Electrical and Space Engineering, Lule\r{a} University of Technology, Lule\r{a} SE-97187, Sweden.}
\address[Second]{WideFind AB, Aurorum 1C, Lule\r{a} SE-97775, Sweden}
%\address[Third]{Division of Engineering and Applied Sciences, California Institute of Technology, Pasadena, California, USA}
\address[Forth]{School of Electronics, Electrical Engineering and Computer Science (EEECS), Queen's University Belfast and Centre for Intelligent Autonomous Manufacturing Systems (i-AMS), United Kingdom}

\begin{abstract}                % Abstract of not more than 250 words.
\glspl{mav} navigation in subterranean environments is gaining attention in the field of aerial robotics, however there are still multiple challenges for collision free navigation in such harsh environments. This article proposes a novel baseline solution for collision free navigation with \gls{nmpc}. In the proposed method, the \gls{mav} is considered as a floating object, where the velocities on the $x$, $y$ axes and the position on altitude are the references for the \gls{nmpc} to navigate along the tunnel, while the \gls{nmpc} avoids the collision by considering kinematics of the obstacles based on measurements from a 2D lidar. Moreover, a novel approach for correcting the heading of the \gls{mav} towards the center of the mine tunnel is proposed, while the efficacy of the suggested framework has been evaluated in multiple field trials in an underground mine in Sweden.
\end{abstract}


\end{frontmatter}
%===============================================================================
\glsresetall
%
%
%
%%%%%%%%%%%%%%%%%%%%%%%%%%%%%%%%%%%%%%%%%%%%%%%%
\section{Introduction}
%%%%%%%%%%%%%%%%%%%%%%%%%%%%%%%%%%%%%%%%%%%%%%%%
Recent technological developments in \glspl{mav} has resulted to a rapid growth of the usage of these platforms in different challenging applications, such as underground mine inspection~\citep{mansouri2020deploying, mansouri2019autonomous}, infrastructure inspection~\cite{mansouri2018cooperative} and subterranean exploration~\citep{rogers2017distributed}. These platforms explore the area and provide access to unreachable, complex, dark and dangerous locations, while minimizing the human involvement in service and inspection procedures.

%leading solution in harsh, unreachable and dangerous environments by 

\begin{figure*}[ht]
\centering
\includegraphics[width=1\linewidth]{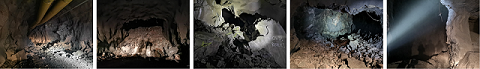}%undergroundmine.png
\caption{Few examples of underground mine environments with limited access and obstacles that shows the need for collision avoidance methods for \gls{mav} navigation.}
\label{fig:minearea}
\end{figure*}

Navigation of aerial robots in underground mine tunnels should cope with multiple challenges as conditions, such as the lack of illumination, narrow passages, crossing paths, limited entrances, dust, high moisture and uneven surfaces directly affect the reliability of sensor measurements and thus the \gls{mav} requires a collision avoidance method. Figure~\ref{fig:minearea} shows photos from one underground mine in Sweden to indicate these conditions in different mine locations. This work addresses the autonomous navigation from a novel perspective by using the concept of a floating object, where the method does not not rely on global information and position estimation, and the dependency on high end and highly expensive sensors is removed. The proposed system, from an application point of view, fits in scenarios towards autonomous underground mine inspection schemes.

This work proposes a low-cost and resource-constrained platform for 
navigation and collision avoidance in underground tunnels. 
The proposed method considers $x$, $y$-velocities and altitude and
attitude estimation, instead of a full pose estimation. 
The \gls{nmpc} is implemented for following the velocities and the
altitude references, while considering the dynamics of the platform 
and avoiding obstacles, based on the measurements from a 2D lidar. 

The \gls{nmpc} optimization problem is solved using OpEn 
\citep{open2019}, which is an open-source solver written in 
Rust: a fast modern programming language, which offers memory 
safety guarantees, an important feature for embedded 
applications. Furthermore, OpEn implements the proximal averaged Newton-type 
method for optimal control \gls{panoc} \citep{sathya2018embedded}, combined with the quadratic penalty method 
\citep{Hermans:IFAC:2018} to accommodate nonlinear constraints.

Moreover, a weighted arithmetic mean based on 2D lidar measurements is 
implemented to correct the heading of the \gls{mav} towards open areas. 
To the best of our knowledge, this is the first work on embedded 
\gls{nmpc} with the concept of floating objects and collision avoidance 
for underground mine navigation. 

\subsection{Background \& Motivation}

\gls{mav} autonomous navigation~\citep{nieuwenhuisen2016layered} with collision avoidance in unknown complex indoor and GPS-denied environments essentially relies on how precise perception of the environment is. There are many works that address this challenging task~\citep{lu2018survey}, mainly focusing on visual and range finding approaches.
%In the related literature, different level of autonomy is defined for the \gls{mav} from position holding up to fully autonomous navigation~\citep{nieuwenhuisen2016layered}. The main essential component for fully autonomous navigation is the functionality to avoid collisions and obstacles successfully. 

Visual or camera-based approaches can provide rich information about the environment, while being relatively lightweight and cheap. There is a lot of research behind \gls{mav} camera-based autonomous navigation. For instance, in~\citep{shen2012stochastic} an algorithm is proposed with collision avoidance, which determines regions for further exploration based on the evolution of a stochastic differential equation. Another approach~\citep{bircher2016receding} for the path generation utilizes a receding horizon path planning algorithm, which acquires visual information from the sensing system and provides a map of the explored environment for collision free navigation and for tracking exploration progress. In~\citep{al2017obstacle} the authors propose an algorithm that uses feature points for obstacle detection and by further estimating their positions via the visual data the \gls{mav} takes the avoidance action. These approaches are based on processing visual information, which reduces map uncertainty and allows obstacle avoidance. However, camera-based approaches typically require well-illuminated environments and high computation power to process visual information for localization, to store the previous position and to calculate the next position, while avoiding revisiting known areas. Additionally the collision avoidance system strongly relies on a map, which can drift or camera, which has limited \gls{fov} and it may result in collision with obstacles. These shortcomings make visual approaches hardly applicable in dark, harsh and large environments.
%There have been many works that addressed the navigation problem with camera-based approaches~\citep{shen2012stochastic, bircher2016receding}, these methods compute regions that reduce map uncertainty, based on current information on the map while avoiding obstacles. These exploration methods evaluated in well-illuminated environments and require in general a high computation power to process the images, to calculate the best next point and to localize accurately and to store the previous information of the map in order to avoid revisiting the area. Moreover, the collision avoidance decouples from the controller and high-level path planner based on the map or \gls{fov} of the camera should avoid collisions and provides feasible paths. However, the map may drift and the camera has limited \gls{fov} and this may result in the collision of the platform. These facts limit the applicability of these methods in harsh and large-scale structures. 

Range finding approaches provide distances to objects in its \gls{fov} and strongly rely on the quality of the metric data. Among a large number of sensors, lidar or laser rangefinder is presently a standard sensor for mobile robotics applications. In the field trials with the \gls{mav} some works address navigation and collision avoidance tasks based on lidars, such as autonomous \gls{mav} inspection inside buildings for maintenance or disaster management~\citep{droeschel2016multilayered}, aerial structure inspection in the outdoor scenario~\citep{azevedo2017collision}, indoor chimney inspection~\citep{quenzel2019autonomous}, etc. All these approaches rely on reactive navigation schemes like potential fields~\cite{kanellakis2018towards} or occupancy grid maps (such as graph-search methods). The main disadvantages of these schemes lie in local optima, limiting their ability to find their way around obstacles without a global planner, and tuning parameters for attractive and repulsive forces for different environments, while the limitation of the map based approaches are the same as discussed above.

%For example, study~\citep{droeschel2016multilayered} shows integrated system which allows autonomous \gls{mav} inspection inside buildings for maintenance or disaster management. Another approach~\citep{azevedo2017collision} shows aerial structure inspection in the outdoor scenario. In~\citep{kanellakis2018towards} authors demonstrate autonomous drone inspection in challenging mine environment. A recent study~\citep{quenzel2019autonomous} shows an aerial platform for indoor chimney inspection. All these approaches rely on reactive navigation schemes like potential field or occupancy grid map. However, the main disadvantage of these schemes are staying in local optimum and tuning parameters for attractive and repulsive forces. The limitation of map based approaches is same as discussed above.
%Additionally, there are works that address the collision avoidance based on lidar range finders~\citep{cowley2011rapid,kanellakis2018towards,droeschel2016multilayered}, these methods rely on reactive navigation schemes as the widely used potential fields or occupancy grid map. The reactive navigation schemes suffers from local optimum and tuning parameters for attractive and repulsive forces. The map based approaches based on lidar face same limitation of map based approaches that discussed above.

There are few works that consider the navigation problem in dark tunnels, such as~\citep{ozaslan2017autonomous}. In this work the authors address estimation, control, navigation and mapping problems for autonomous tunnel inspection using \gls{mav}s validating their approach with field trials. However, the use of a high-end sensor suit limits the applicability of the overall method. In~\citep{bharadwaj2016small} the authors proposed an infrared camera for navigation in dark environments, however the work doesn't provide autonomous navigation of the \gls{mav}. In~\citep{dang2019field} the approach for autonomous aerial navigation in a subterranean environment is proposed, while the method is evaluated in underground tunnels. The approach relied on the global planner and a high end sensor for collision avoidance. This results to collision in case of drifts in the map and the \gls{mav} should have a low-level collision avoidance scheme. In~\citep{huang2019duckiefloat} autonomous blimp Duckiefloat is proposed for tunnel navigation, however collision avoidance is not considered due to non-rigid body of the platform and it is collision-tolerant.

% there are also works~\citep{bharadwaj2016small, hennage2019fully} that use infrared camera for navigation in dark environments but they still require future evaluation of the developed algorithms.

%There have been very few works that consider the navigation problem in dark tunnels such as~\citep{ozaslan2017autonomous}. The authors addressed the problem of estimation, control, navigation and mapping problems for autonomous inspection of tunnels using aerial vehicles with the overall approach to be validated through field trials, however, the high-end sensor suit that was utilized for the \gls{mav} has limited the applicability of the overall method. In~\cite{}

\subsection{Contributions}
The first and major contribution of this article is the development of the \gls{mav} as a floating object, while not relying on full odometry information. As position information drifts over time, especially in harsh environments, such as mines with dust, low illumination, wet surfaces, etc. In the worse case scenario, this may result in failure of the mission or collision of the platform. The proposed method can be used as a baseline controller that guarantees collision free navigation based on local information, while a higher level controller can be used for generating reference velocities based on a global map and exploration requirements.

The second contribution is the development and experimental 
validation of a lidar-based navigation system with collision 
avoidance capabilities using \gls{nmpc}. The associated 
optimization problem is solved on an Aaeon UP-Board embedded platform 
onboard the MAV, using Rust code that is auto-generated using OpEn
\citep{open2019}.

The third contribution specifically highlights the heading command generation as the \gls{mav} has to identify the direction of the flight in the tunnel navigation task. To address this issue, this work proposes a geometry-based approach to obtain the heading rate towards the open area, which is improved by integrating the gyro response for better robustness.

Finally, the fourth major contribution, stems from evaluating the performance of the proposed method in an underground mine located in Sweden. As it is presented the proposed method has a significant novelty and impact on embedded \gls{nmpc} navigation with collision avoidance for underground tunnels. The following link \underline{https://youtu.be/-MP4Sn6Q1uo} provides a video summary of the overall results.  

\subsection{Outline}
The rest of the article is structured as follows. Initially the preliminaries are explained in Section~\ref{sec:preliminaries}, then the mathematical models of the \gls{mav} and the obstacles are described in Section~\ref{sec:model}. Next, a presentation of the \gls{nmpc} formulation and the solver is described in Section~\ref{sec:NMPC}. The geometry approach for generating heading rate commands is discussed in Section~\ref{sec:heading}. Section~\ref{sec:result} presents the hardware setup and provides the experimental results from the underground mine with uneven surfaces. Finally, Section~\ref{sec:Conclusions} concludes the article by summarizing the findings and offering some directions for future research.
%%%%%%%%%%%%%%%%%%%
\section{Preliminaries} \label{sec:preliminaries}
%%%%%%%%%%%%%%%%%%%
In this article, the main part to enable autonomous navigation is to consider the \gls{mav} as floating object, to remove dependencies on accurate localization schemes. Thus, the position estimation in $x$ and $y$ axes is not considered. The \gls{mav} is equipped with multiple sensor suits, such as on-board \gls{imu} to estimate the attitude $[\phi,\theta]^\top$, an optical flow sensor for linear velocities $[v_x,v_y]^\top$, and one beam lidar for height $z$ and $v_z$ velocity estimation. Moreover, the 2D lidar provides a set of $R$ ranges with $r$ to denote the range generated at an angle of rotation $\xi$, with the angular resolution depend on the 2D lidar characteristics. The one beam lidar provides the distance to the ceiling $d^{+z}$. The main challenge of navigation is to avoid collision and extract a proper heading to follow an obstacle free path along the tunnels, while the position of the \gls{mav} is not known. This article proposes the novel approach and considers the kinematic of the obstacles in respect to the \gls{mav}, while no global information of obstacle positions is available. 
%%%%%%%%%%%%%%%%%%%
\section{MAV Dynamics and Obstacle Kinematics} \label{sec:model}
\begin{figure} [htbp!] \centering
\includegraphics[width=0.8\linewidth]{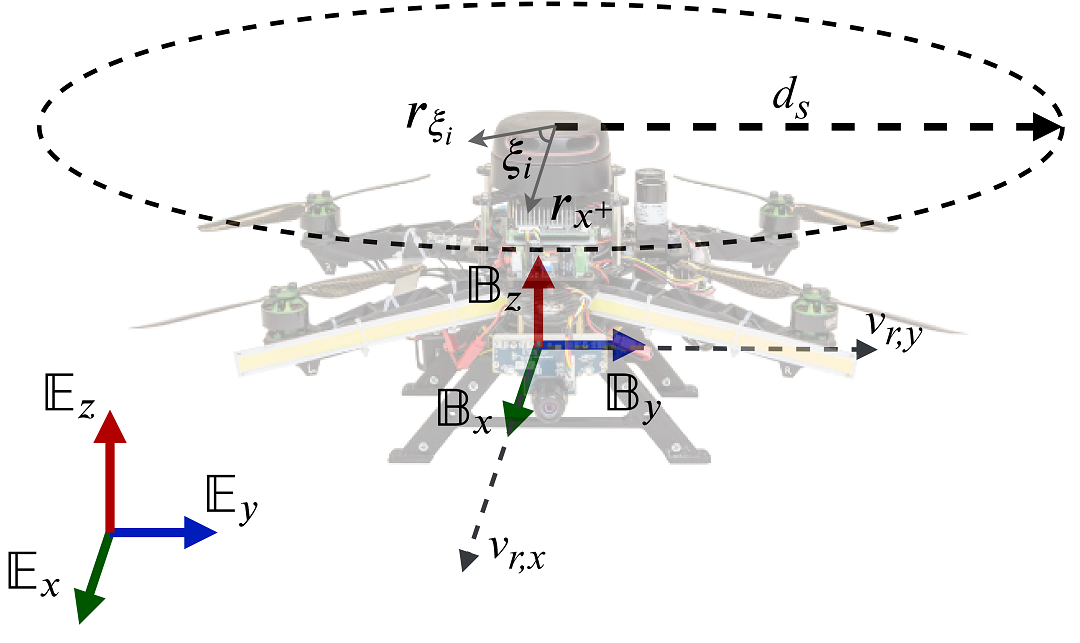}%undergroundmine.png
\caption{Illustration of the \gls{mav} with the attached body fixed frame $\mathbb{B}$ and inertial frame $\mathbb{E}$, while the safety distance is shown by $d_s$, reference velocities by $v_{r,x}$, and $v_{r,y}$, lidar ranges by $r$ and $\xi$ is the angle correspond to the range.}
\label{fig:coordinateframes}
\end{figure}

%%%%%%%%%%%%%%%%%%%
The \gls{mav} is considered as a six \gls{dof} object with a Body-Fixed Frame $\mathbb{B}$ attached and the inertial frame $\mathbb{E}$ as depicted in Figure~\ref{fig:coordinateframes}. 
The \gls{mav} is modelled by the position of the center of the mass in the inertia frame and the orientation of the body around each axes with respect to the inertial frame~\citep{mav_linear_mpc}. This modelling requires a full state estimation for the position and orientation in the inertial frame. However, underground mines are \gls{gps}-denied environments, \gls{imu} measurements are affected by accumulated errors and due to low illumination and lack of features, the performance of the vision based localization methods, such as \gls{vo}~\citep{balamurugan2016survey} is limited. Thus, in this article the \gls{mav} dynamics is defined in the body frame and does not depend on position estimation of the $x$ and $y$-axes, to avoid affecting the performance of the controller by drifts and uncertainties in $x$, $y$ positions. These drifts do not affect the proposed robust controller for collision avoidance either, while the \gls{mav} is modelled by \eqref{eq:modeluav} in body frame as:
\begin{subequations} \label{eq:modeluav}
\begin{align}
        \dot{z}(t) &= v_z(t),  \\ 
        \dot{v}(t) &= R_{x,y}(\theta,\phi) 
        \begin{bmatrix} 0 \\ 0 \\ T \end{bmatrix} + 
        \begin{bmatrix} 0 \\ 0 \\ -g \end{bmatrix} - 
        \begin{bmatrix} A_x & 0 & 0 \\ 0 &  A_y & 0 \\ 0 & 0 & A_z \end{bmatrix} v(t),   \\
        \dot{\phi}(t) & = \nicefrac{1}{\tau_\phi} (K_\phi\phi_d(t)-\phi(t)),  \\
        \dot{\theta}(t) & = \nicefrac{1}{\tau_\theta} (K_\theta\theta_d(t)-\theta(t)),
\end{align}
\end{subequations}
% <-- THIS SKIPS A LINE BREAK (DO NOT REMOVE)
where $v = (v_x, v_y, v_z) \in \mathbb{R}^3$ is the vector of linear velocities, 
$\phi \in \mathbb{R} \cap [-\pi/2,\pi/2]$ and $\theta \in \mathbb{R}\cap [-\pi/2,\pi/2]$ are the roll and pitch angles, $R_{x,y}$ is the rotation matrix about 
the $x$ and $y$ axes, $T \in [0,1]$ is the mass-normalized thrust, $g$ is the gravitational acceleration, 
$A_x\in \mathbb{R}$, $A_y\in \mathbb{R}$, and $A_z\in \mathbb{R}$ are the normalized mass drag coefficients
The low-level control system is approximated by first-order dynamics driven by the desired pitch and roll angles $\phi_d$ and $\theta_d$ with gains of $K_\phi, K_\theta\in\mathbb{R}^2$.

Additionally, five obstacles are considered for $x^+$, $x^-$, $y^+$, $y^-$, $z^+$ axes for collision avoidance in regards to the surroundings. These obstacles are modeled as surfaces perpendicular to the corresponding axes, 
while the distance of the surface $d$ is the minimum range extracted from 
2D lidar ranges in corresponding axes. The obstacle kinematics is presented by \eqref{eq:obstkinematic} in the body frame, where $\dot{d}$ is the velocity of the surface. As an example, the obstacle in the $x$ axes has the $-v_x$ velocity compared to the \gls{mav} velocity in the body frame. In this approach, the distance to the obstacles can be predicted based on the \gls{mav} maneuvers.
\begin{subequations} \label{eq:obstkinematic}
\begin{align}%\mathrm{obs},1
       \dot{d}^{x^+}(t) &= -v_x(t),   \\ 
       \dot{d}^{x^-}(t) &= v_x(t),  \\ 
       \dot{d}^{y^+}(t) &= -v_y(t), \\ 
       \dot{d}^{y^-}(t) &= v_y(t),  \\ 
       \dot{d}^{z^+}(t) &= -v_z(t). 
\end{align}
\end{subequations}
%%%%%%%%%%%%%%%%%%%
\section{Nonlinear Model Predictive Control} \label{sec:NMPC}
%%%%%%%%%%%%%%%%%%%

\begin{figure*}[htbp!] \centering 
\input{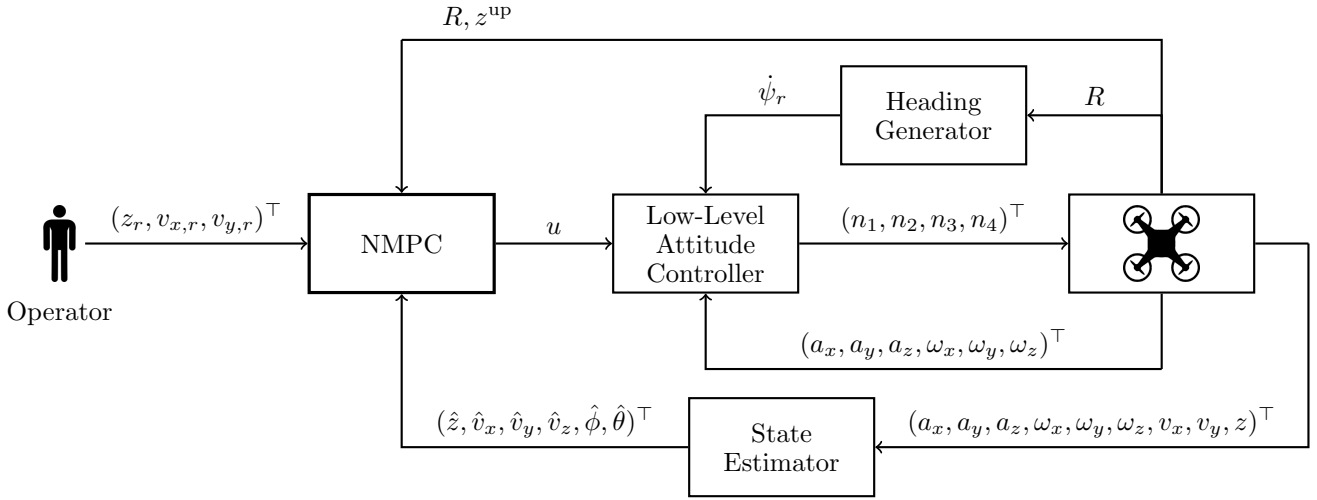}
\caption{Control scheme of the proposed navigation module, where the heading commands are provided from the geometry approach. The \gls{nmpc} generates thrust and attitude commands, while the low level controller generates motor commands $[n_1,\dots,n_4]^\top$. The velocity estimation is based on \gls{imu} measurements, optical flow and downwards facing single beam lidar.}%\vspace{-1cm}
\label{fig:Controllerscheme}
\end{figure*}

In this article, the \gls{nmpc} with collision avoidance constraints are developed and solved by \gls{panoc}~\citep{sathya2018embedded} to guarantee real-time performance. The \gls{nmpc} objective is to generate attitude $\phi_d, \theta_d$ and thrust commands $T_d$ for the low-level controller, while the reference altitude and velocities $[z_r, v_{r,x},v_{r,y}]^\top$ are provided from the operator. The \gls{pid} controller in the flight controller generates the motor commands $[n_1,\dots,n_4]^\top$ for the \gls{mav}. A block diagram representation of the proposed \gls{nmpc} and the corresponding low-level controller is shown in Figure~\ref{fig:Controllerscheme}.

  %\vspace{-0.1cm}

Based on the non-linear dynamics of the \gls{mav} \eqref{eq:modeluav} and kinematics of the obstacles \eqref{eq:obstkinematic}, 
the state of the system is 
\(
	  \bm{x}
    {}={}
	  [z,v_x,v_y,v_z,\phi,\theta,d^{x^+},d^{x^-},d^{y^+},d^{y^-},d^{z^+}]^\top
\), 
\(
	  \hat{\bm{x}}
    {}={}
	  [\hat{z},\hat{v_x},\hat{v_y},\hat{v_z},\hat{\phi},\hat{\theta}]^\top
\) 
is the estimated state from \gls{ekf} for \gls{mav} dynamics, while the distances to the surfaces are obtained from the 2D lidar range measurement, and the control input is: 
\(
	  u
    {}={}
	  [\phi_d,\theta_d,T]^\top
\). 
By discretizing \eqref{eq:modeluav} with the Euler method and with a sampling time of $T_s$, we obtain a discrete-time dynamical system as: 
\[
 \bm{x}_{t+1} = f(\bm{x}_t, u_t).
\]

In the \gls{nmpc} approach, at every time instant $k$, a finite-horizon problem with prediction horizon $N \in \mathbb{N}^{\ge 2}$ is solved, while $\bm{x}_{k+j|k}$, and $u_{k+j\mid k}$ are the states and control actions $k+j$ steps ahead of the current time step $k$. \gls{nmpc} generates an optimal sequence of control actions $u_{k|k}^{\star}$, $\dots$, $u_{k+N-1|k}^{\star}$ and the first control action $u_{k|k}^\star$ is applied to the flight controller using a zero-order hold element, that is, $u_t=u_{k|k}^\star$ for $t\in [kT_s, (k+1)T_s]$. For the proposed \gls{nmpc}, the following finite horizon cost function is introduced:
\begin{multline} \label{eq:costfunction} 
J = \sum_{j=0}^{N-1} 
  \underbrace{\|z_{k+j+1|k}-z_{r}\|_{Q_z}^2}_\text{altitude error}  
   + \underbrace{\|v_{k+j+1|k}-v_{r}\|_{Q_v}^2}_\text{velocity error}\\   
   + \underbrace{\|u_{k+j+1|k}-u_{r}\|_{Q_u}^2}_\text{actuation}
   + \underbrace{\|u_{k+j|k}-u_{k+j-1|k}\|_{Q_{\Delta u}}^2 }_\text{smoothness cost}.
\end{multline} 

The first and second terms of $J$ are the tracking of the desired altitude, and velocity references respectively. 
The third term is the hovering term, where $u_{ref}$ is $[0,\, 0,\, g]^\top$, which is the 
hover thrust with horizontal angles. 
The forth term penalizes the aggressiveness of the obtained control actions. 
Additionally, $Q_z \in \mathbb{R}$, $Q_v \in \mathbb{R}^{3\times3}$, $Q_u\in \mathbb{R}^{3\times 3}$, 
$Q_{\Delta u}\in \mathbb{R}^{3\times 3}$ are the weights for each term of the objective function, which reflects the relative importance of each term.

Additionally, to avoid the obstacles, five constraints are defined in~\eqref{eq:constraints}, 
while the \gls{nmpc} predicts the distance of the obstacles based on the kinematic of the constraints \eqref{eq:obstkinematic}
\begin{subequations} \label{eq:constraints}
\begin{align} 
&  d_{s} - d^{x^+}_{k+j\mid k} - \dot{d}^{x^+}_{k+j+1\mid k}T_s \le 0,   \label{eq:constraints:a}\\
&  d_{s} - d^{x^-}_{k+j\mid k} - \dot{d}^{x^-}_{k+j+1\mid k}T_s \le 0, \\
&  d_{s} -d^{y^+}_{k+j\mid k} - \dot{d}^{y^+}_{k+j+1\mid k}T_s \le 0, \\
&  d_{s} - d^{y^-}_{k+j\mid k} - \dot{d}^{y^-}_{k+j+1\mid k}T_s \le 0, \\
&  d_{s} -d^{z^+}_{k+j\mid k} - \dot{d}^{z^+}_{k+j+1\mid k}T_s  \le 0, 
\end{align}
\end{subequations}
% <-- THIS SKIPS A LINE BREAK (DO NOT REMOVE)
where $d_s$ is the defined safety distance given by the operator. 
The proposed constraints guarantee that the \gls{mav} has at least $d_s$ distance to each obstacle. The following optimization problem can be defined:
\begin{subequations} \label{eq:nmpc}
\begin{align}
&\underset{
  \{  u_{k+j\mid k} \}_{j=0}^{N-1}} {\mathbf{minimize}}\  J \\
&\mathbf{subject\,to}:\bm{x}_{k+j+1\mid k}=f(\bm{x}_{k+j\mid k},u_{k+j\mid k}), 
\label{eq:optimizationcoverage:sysDyn}
\\
& \text{Constraints~\eqref{eq:constraints}},\label{eq:optimizationcoverage:constraint}
\\ 
&u_{k+j\mid k} \in [u_{\min},u_{\max}],
\label{eq:optimizationcoverage:actuationConstraints}
\end{align}
\end{subequations}
for $j=0,\ldots, N-1$.

Problem \eqref{eq:nmpc} is a parametric non-convex optimization problem that must be solved online, by typical on-board computation platforms with limited computing resources and within the hard run-time requirements of the navigation controller. 
To that end, we use the fast and efficient optimization solver Optimization Engine (for short 
\texttt{OpEn}) developed by~\cite{open2019}. \texttt{OpEn} can solve problems of the general form 
\begin{subequations}\label{eq:open_problem_format}
\begin{align}
 \mathbf{Minimize}_{u \in U}\,& \ell(u),
 \\
 \mathbf{subject\,to}:\,& F(u) = 0,
\end{align}
\end{subequations}
% <-- THIS SKIPS A LINE BREAK (DO NOT REMOVE)
by means of the penalty method and PANOC. The solver allows the designer to provide the 
problem specification in Python and generates code written in Rust (\texttt{https://rust-lang.org}): 
a fast programming language that guarantees safe memory management. This is an important
desideratum for safety-critical applications such as \gls{mav} navigation.
In order to write Problem \eqref{eq:nmpc} in the form of \eqref{eq:open_problem_format}, we first
eliminate the state sequence following the procedure detailed in \citep{sathya2018embedded}.
This way, we have a problem of the decision variable $u=(u_{k\mid k}, u_{k+1\mid k},\ldots, , u_{k+N-1\mid k})\in\mathbb{R}^{3N}$,
which is constrained in $U=[u_{\min}, u_{\max}]^{N}$.
Constraint \eqref{eq:constraints:a} is equivalent to 
\begin{equation}
 \max\left\{0, d_{s} - d^{x^+}_{k+j\mid k} - \dot{d}^{x^+}_{k+j+1\mid k}T_s\right\} = 0.
\end{equation}
Likewise, constraints \eqref{eq:constraints} can all be written as equality constraints.

%%%%%%%%%%%%%%%%%%%
\section{Heading Correction} \label{sec:heading}
%%%%%%%%%%%%%%%%%%%
The purpose of this module is to guide the \gls{mav} towards the direction that has the largest amount of navigable space. The 2D lidar provides range measurements for each degree of of rotation, thus the weighted arithmetic mean~\citep{madansky2017weighted} is proposed. In the weighted arithmetic mean, the range value for each angle is used as a weight for the corresponding angle. The weighted arithmetic mean is calculated as:
\begin{equation} \label{eq:heading}
    \psi_{k}=\frac{\sum_{i}r_{i,k}\xi_i}{\sum_{i} r_{i,k}}
\end{equation}
where $i \in [i_{min}, ~i_{max}]$ is the set of beam angle indexes and $r_i$ is the range measurement for each corresponding angle $\xi$ of 2D lidar rotation. $\xi_{min}$ and $\xi_{max}$ define a range of angles, as shown in the Figure~\ref{fig:coordinateframes}, and there are angles for $y^+$ and $y^-$ axes in the body frame, thus the range measurements from $y^-$ axis up to $y^+$, based on the right hand rule are used. However, this result is noisy and subject to spurious changes in the environment. For improving this, the integrated $z$-axis of the gyro in the \gls{imu} is used as a prior, using a complementary filter as follows.

Initially, the heading is integrated from the $z$-axis angular rate to predict the movement as:
\begin{equation} \label{eq:heading_est}
     \hat{\psi}_{k|k-1} = \hat{\psi}_{k-1|k-1} + \omega_{z,k} T_s,
\end{equation}
which in the sequel is corrected using the estimated angle from the 2D lidar: 
\begin{equation} \label{eq:heading_est_2}
     \hat{\psi}_{k|k} = \beta \hat{\psi}_{k|k-1} - (1 - \beta)\psi_k,
\end{equation}
where $\beta \in (0,~1)$ is a classic complementary filter, which was chosen due to its simplicity in tuning for generating robust results. It should be also noted the negation in the second term, $ - (1 - \beta)\psi_k$, that is generated from the fact that the measured free space always rotates with the inverted angular rates relative to the \gls{mav}.

The obtained heading angle is fed to the low level proportional controller with gain of $k_p > 0$ to always have the \gls{mav} point toward the open space:
\begin{equation}
    \label{eq:headingrate}
    \dot{\psi}_{r,k} = -k_p \hat{\psi}_{k|k}.
\end{equation}

%%%%%%%%%%%%%%%%%%%%%%%%%%%%%%%%%%%%%%%%%%%%%%%%\caption{Control scheme of the proposed navigation module, while the heading commands is provided from the geometry approach. The \gls{nmpc} generates thrust and attitude commands, while the low level controller generates motor commands $[n_1,\dots,n_4]^\top$. The velocity estimation is based on \gls{imu} measurements, optical flow and single beam lidar looking downward. }

\section{Results} \label{sec:result}
%%%%%%%%%%%%%%%%%%%%%%%%%%%%%%%%%%%%%%%%%%%%%%%%
This section describes the experimental setup and the experimental evaluation that has been performed in an underground mine.
\newline
Link: \underline{https://youtu.be/-MP4Sn6Q1uo} provides a video summary of the system.  

\subsection{Experimental Setup}
\begin{figure*}[!htb]
    \centering
\includegraphics[width=\linewidth]{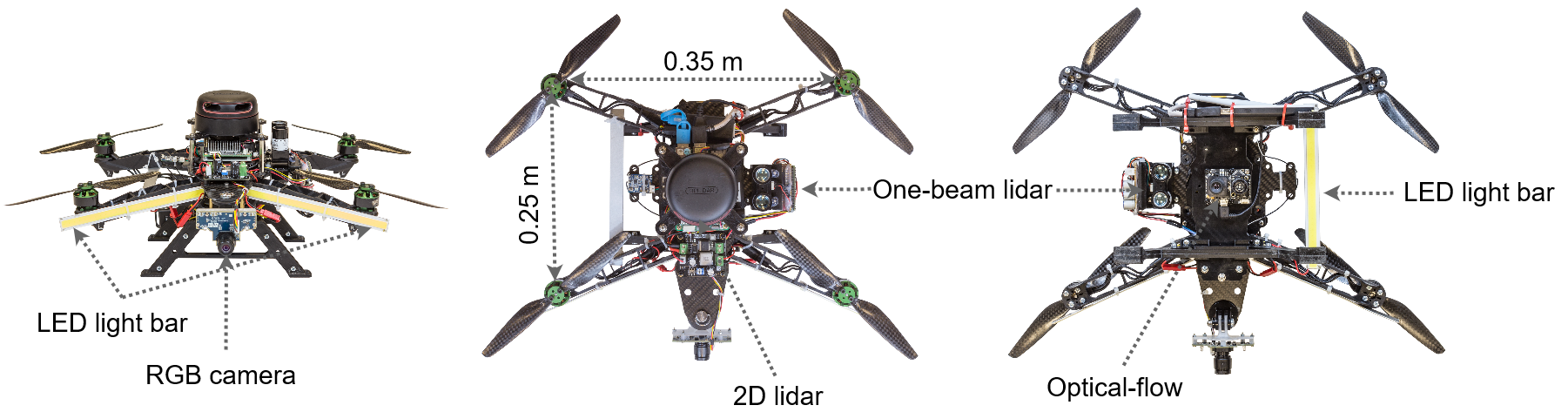}
    \caption{The developed quad-copter equipped with a forward looking camera, a LED lights, optical flow, 2D lidar and single beam lidar looking upward and downward.}
    \label{fig:LTUplatfrom}
\end{figure*}

In this work, a low-cost quad-copter that is developed at Lule{\aa} University of Technology has been operated in an underground mine. The developed platform is presented in~\citep{mansouri2020deploying}, while the platform is equipped with an additional upwards facing single beam Lidar-lite v3 at $\unit[100]{hz}$ for measuring distance to the ceiling. Figure~\ref{fig:LTUplatfrom} depicts the platform highlighting the sensor suits and dimensions.

\subsection{Experimental Evaluation}
This work evaluates the method in two different types of environments; a tunnel environment and a tunnel blockage. This highlights the adaptability and feasibility of the methods in different scenarios in the underground mines. Two locations are selected for this purpose, the first location is an underground iron mine with strong magnetic fields, while the second one is underground tunnels without corrupting magnetic fields, both environments are located in Sweden. The first location has a width and height of approximately $\unit[6]{m}$ and $\unit[4]{m}$ with tunnels and blockage areas. The second location morphology resembled an \textit{S} shape and the dimensions of the area, where the \gls{mav} navigates autonomously were $3.5 (\text{width}) \times 3 (\text{height})~\unit{m^2}$. It should be highlighted that none of the flights resulted to a collision and the flights were terminated only due to battery drainage.

The tuning parameters of the \gls{nmpc} and the parameters of the \gls{mav} model are presented in the Tables~\ref{table:nmpcparameters} and~\ref{table:mavparameter} correspondingly. The \gls{nmpc} prediction horizon $N$ is 40 and the control sampling time $T_s$ is $\unit[0.05]{s}$. The $\beta$ and $k_p$ for the heading generator approach are 0.95 and 0.03 respectively. During the navigation task the average solver time for the optimization framework is around 10$\unit[]{ms}$, while it consumes an average of 10\% of the CPU usage. 

{\renewcommand{\arraystretch}{1.3}
\begin{table}[htbp!]
\centering
\caption{The tuning parameters of the \gls{nmpc}.}
\label{table:nmpcparameters}
\begin{tabular}{ccccc}
\hline
\multicolumn{1}{|c|}{$Q_z$}    & \multicolumn{1}{c|}{$Q_v$}     & \multicolumn{1}{c|}{$Q_u$}        & \multicolumn{1}{c|}{$Q_{\Delta u}$}       & \multicolumn{1}{c|}{$d_{s}$}    \\ \hline
\multicolumn{1}{|c|}{10}   & \multicolumn{1}{c|}{$[5,5,5]^\top$} & \multicolumn{1}{c|}{$[20,20,20]^\top$} & \multicolumn{1}{c|}{$[20,20,20]^\top$} & \multicolumn{1}{c|}{$\unit[1]{m}$}         \\ \hline
                           &                            &                               &                               &                                \\ \hline
\multicolumn{1}{|c|}{$T$} & \multicolumn{1}{c|}{$\phi_{min}$}  & \multicolumn{1}{c|}{$\phi_{max}$}  & \multicolumn{1}{c|}{$\theta_{min}$}   & \multicolumn{1}{c|}{$\theta_{max}$} \\ \hline
\multicolumn{1}{|c|}{$[0,1]\cap \mathbb{R}$}    & \multicolumn{1}{c|}{$\unit[-0.4]{rad/s}$}     & \multicolumn{1}{c|}{$\unit[0.4]{rad/s}$}        & \multicolumn{1}{c|}{$\unit[-0.4]{rad/s}$}        & \multicolumn{1}{c|}{$\unit[0.4]{rad/s}$}        \\ \hline
\end{tabular}
\end{table}
}
{\renewcommand{\arraystretch}{1.3}
\begin{table}[htbp!]
\centering
\caption{The parameters of the \gls{mav} model.}
\label{table:mavparameter}
\begin{tabular}{|cccccccc|}
\hline
$g$    & $A_x$  & $A_y$  & $A_z$  & $K_\phi$  & $K_\theta$  & $\tau_\phi$  & $\tau_\theta$ \\ \hline
$\unit[9.8]{m/{s}^2}$ & 0.1 & 0.1 & 0.2 & 1  & 1 & $\unit[0.5]{s}$ & $\unit[0.5]{s}$ \\ \hline
\end{tabular}
\end{table}
}

\subsubsection{Tunnel Environment:}
\begin{figure}[htbp!]
\centering
\includegraphics[width=\linewidth]{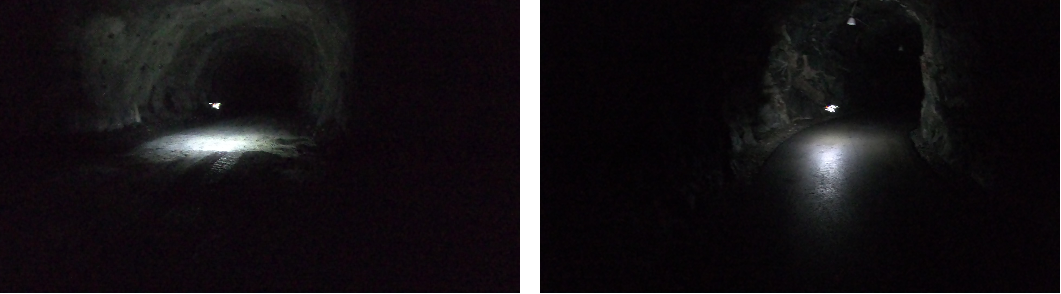}
        \caption{The tunnel environments, while the first and second locations are left and right pictures respectively.}
       \label{fig:resulttunnel}
\end{figure}

The \gls{mav} is evaluated in the two tunnels with different dimensions. Figure~\ref{fig:resulttunnel} shows the tunnel environments, while the \gls{mav} performs autonomous flight with on-board lights.

\begin{figure}[htbp!]
\setlength\fwidth{0.8\linewidth}
\centering
        \input{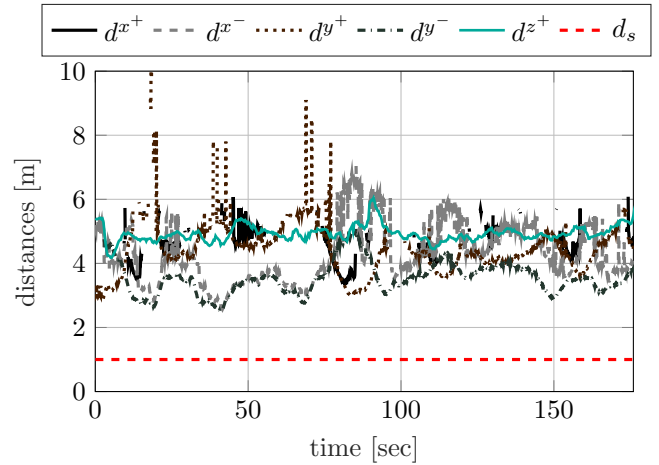}
        \caption{The distance to the obstacles in $x^+$, $x^-$, $y^+$, $y^-$, $z^+$ 
                 axes of the \gls{mav}, while the safety distance is 
                 set to $\unit[1]{m}$ in the first tunnel environment.}
       \label{fig:distances2019-09-12-08-59-03}
\end{figure}

\begin{figure}[htbp!]
\setlength\fwidth{0.8\linewidth}
\centering
        \input{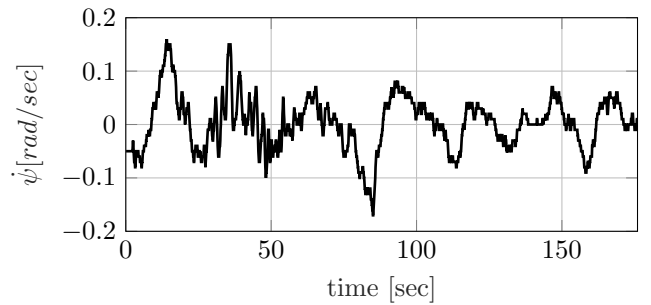}
        \caption{The heading rate command from the weighted arithmetic mean approach in the first tunnel environment, while the \gls{nmpc} avoids obstacles.}
       \label{fig:headrate2019-09-12-08-59-03}
\end{figure}

Figure~\ref{fig:distances2019-09-12-08-59-03} shows the distances from the surroundings of the \gls{mav} in the first tunnel environment, where the $d^{x^+}$, $d^{x^-}$, $d^{y^+}$, $d^{y^-}$, $d^{z^+}$ are distances to the corresponding body frame axes. In this test, the \gls{mav} autonomously navigates with $z_{r}=\unit[1.0]{m}$, $v_{x,r}=\unitfrac[0.5]{m}{s}$, $v_{y,r}=\unitfrac[0.0]{m}{s}$. The maximum range of the used 2D lidar is $\unit[15]{m}$, however it reports $\inf$ for ranges out of this bound or if the laser beam does not bounce back due to wet surfaces. In the following figures the $\inf$ values are not shown and it results to discontinuously in distance plots.
Moreover, Figure~\ref{fig:headrate2019-09-12-08-59-03} provides the heading rate generated, which guides the \gls{mav} towards open spaces. 

\begin{figure}[htbp!]
\setlength\fwidth{0.8\linewidth}
\centering
        \input{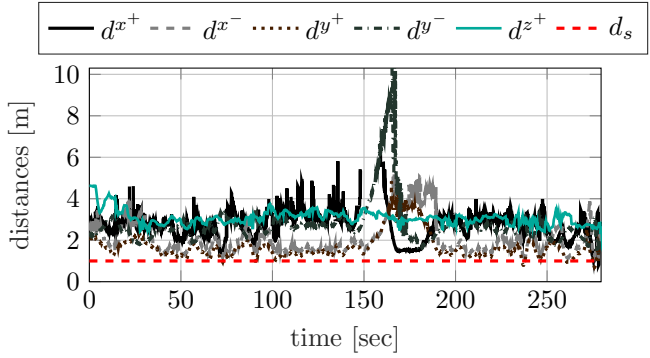}
        \caption{The distance to the obstacles in $x^+$, $x^-$, $y^+$, $y^-$, $z^+$ axes of the \gls{mav}, while the safety distance is set to $\unit[1]{m}$ in the second tunnel environment.}
       \label{fig:distances2019-09-04-08-57-25}
\end{figure}

\begin{figure}[htbp!]
\setlength\fwidth{0.8\linewidth}
\centering
        \input{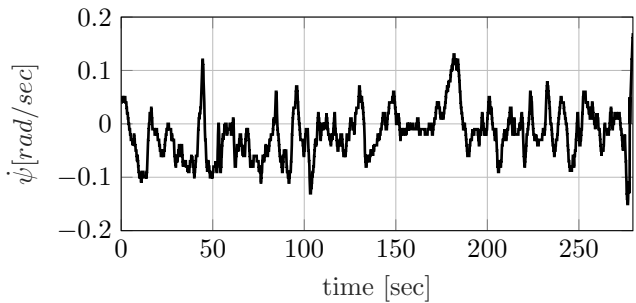}
        \caption{The heading rate command from a weighted arithmetic mean approach in the case of second tunnel environment navigation, while the \gls{nmpc} avoids obstacles.}
       \label{fig:headrate2019-09-04-08-57-25}
\end{figure}

In the second tunnel environment, the \gls{mav} autonomously navigates with $z_{r}=\unit[1.0]{m}$, $v_{x,r}=\unitfrac[1.2]{m}{s}$, $v_{y,r}=\unitfrac[0.0]{m}{s}$, the Figure~\ref{fig:distances2019-09-04-08-57-25} shows the distances from the surrounding of the \gls{mav}. The dimensions of the second tunnel are smaller, when compared to the first tunnel, however the controller still guarantees a safety distance from all the surface obstacles.
Additionally, the generated heading rate command is depicted in Figure~\ref{fig:headrate2019-09-04-08-57-25}.

\subsubsection{Tunnel Blockage Environment:}
\begin{figure}[htbp!]
\centering
\includegraphics[width=\linewidth]{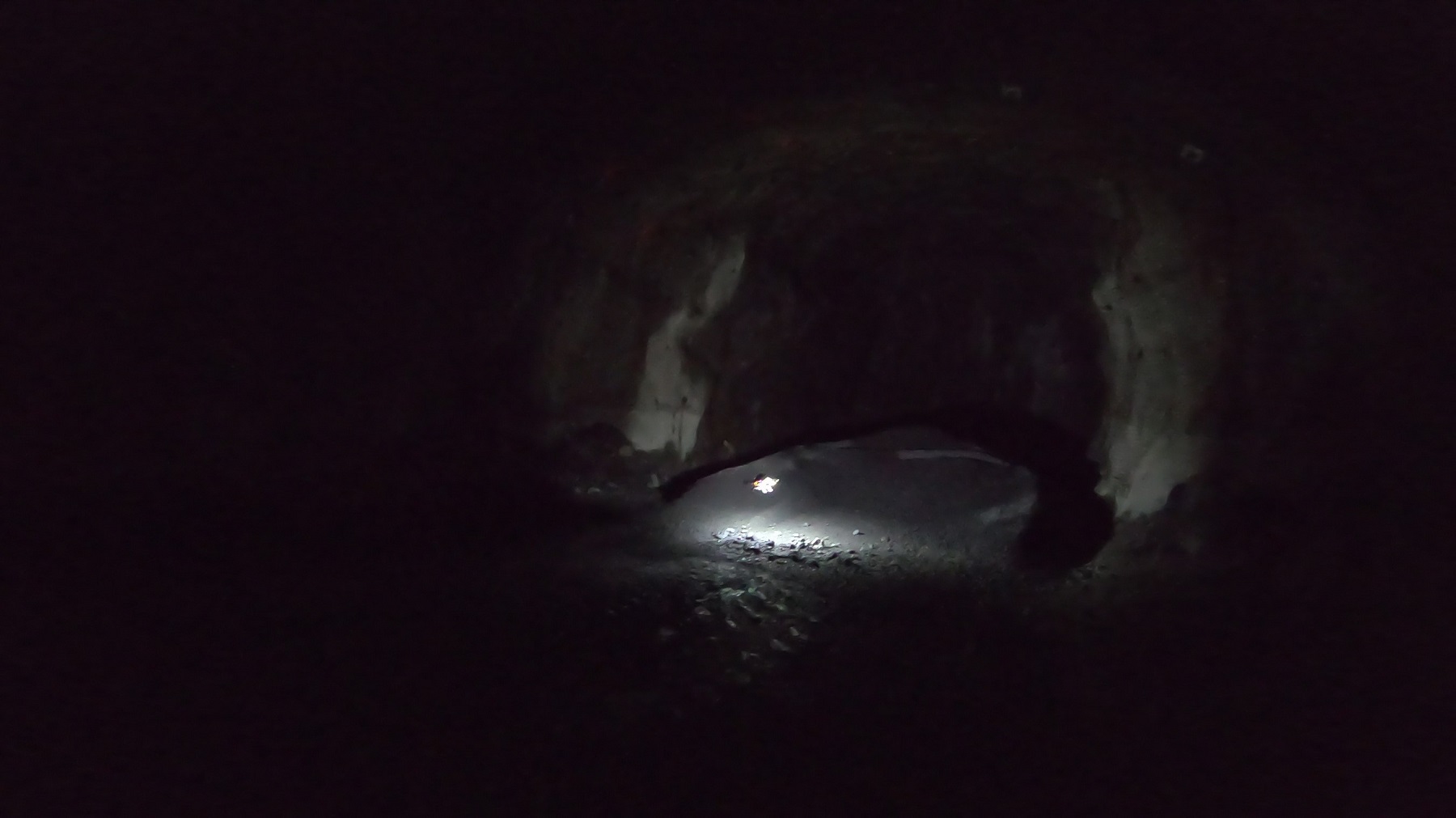}
        \caption{The tunnel blockage environments, while the \gls{mav} navigates autonomously and avoid obstacles.}
       \label{fig:resultblockage}
\end{figure}

In this case the \gls{mav} navigates through the blockage area, while the constant references of $z_{r}=\unit[1.0]{m}$, $v_{x,r}=\unitfrac[0.5]{m}{s}$, $v_{y,r}=\unitfrac[0.0]{m}{s}$ are fed to the \gls{nmpc}. When the \gls{mav} reaches the end of the tunnel, the return command is transmitted and the \gls{mav} returns to the starting point and passes the blockage again. This test shows the applicability of the method in applications regarding underground mines to navigate in blocked areas and return to the base for the reports. Figure~\ref{fig:resultblockage} depicts the tunnel blockage environment, where the \gls{mav} navigation is evaluated, the altitude reference for the \gls{mav} was constant during the experiment, however the \gls{mav} passed the blockage successfully.

\begin{figure}[htbp!]
\setlength\fwidth{0.8\linewidth}
\centering
        \input{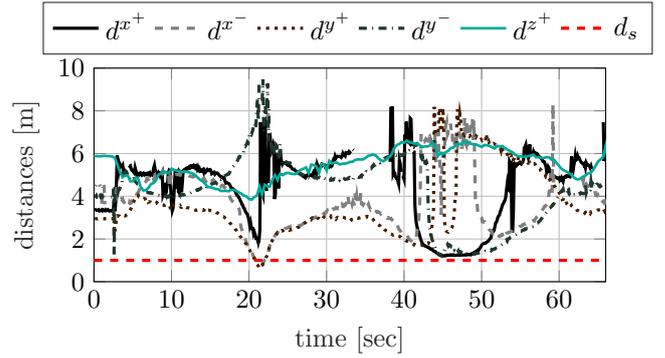}
        \caption{The distance to the obstacles in $x^+$, $x^-$, $y^+$, $y^-$, $z^+$ axes of the \gls{mav}, while the safety distance is set to $\unit[1]{m}$ in the tunnel blockage environment.}
       \label{fig:distances2019-09-11-13-01-57}
\end{figure}

Figure~\ref{fig:distances2019-09-11-13-01-57} depicts the distance to obstacles in each axes of the \gls{mav}, while the \gls{nmpc} guarantees safety distance in all directions. 

\begin{figure}[htbp!]
\setlength\fwidth{0.8\linewidth}
\centering
\begin{tikzpicture}

\begin{axis}[%
width=0.951\fwidth,
height=0.4\fwidth,
at={(0\fwidth,1.1\fwidth)},
scale only axis,
xmin=0.00,
xmax=66.00,
xlabel style={font=\color{white!15!black}},
xlabel={time [sec]},
ymin=-3.20,
ymax=0.40,
ylabel style={font=\color{white!15!black}},
ylabel={$\dot{\psi} [rad/sec]$},
axis background/.style={fill=white},
xmajorgrids,
ymajorgrids,
legend style={legend cell align=left, align=left, draw=white!15!black}
]
\addplot [color=black, line width=1.0pt]
  table[row sep=crcr]{%
0.10	-0.71\\
0.20	-0.71\\
0.30	-0.71\\
0.40	-0.71\\
0.50	-0.71\\
0.60	-0.71\\
0.70	-0.71\\
0.80	-0.71\\
0.90	-0.71\\
1.00	-0.71\\
1.10	-0.71\\
1.20	-0.71\\
1.30	-0.71\\
1.40	-0.71\\
1.50	-0.71\\
1.60	-0.71\\
1.70	-0.71\\
1.80	-0.71\\
1.90	-0.71\\
2.00	-0.71\\
2.10	-0.71\\
2.20	-0.71\\
2.30	-0.71\\
2.40	-0.69\\
2.50	-0.60\\
2.60	-0.48\\
2.70	-0.37\\
2.80	-0.26\\
2.90	-0.18\\
3.00	-0.12\\
3.10	-0.08\\
3.20	-0.05\\
3.30	-0.02\\
3.40	0.00\\
3.50	0.02\\
3.60	0.03\\
3.70	0.03\\
3.80	0.03\\
3.90	0.04\\
4.00	0.04\\
4.10	0.04\\
4.20	0.04\\
4.30	0.04\\
4.40	0.04\\
4.50	0.05\\
4.60	0.05\\
4.70	0.05\\
4.80	0.05\\
4.90	0.05\\
5.00	0.05\\
5.10	0.06\\
5.20	0.06\\
5.30	0.06\\
5.40	0.06\\
5.50	0.06\\
5.60	0.06\\
5.70	0.07\\
5.80	0.07\\
5.90	0.07\\
6.00	0.07\\
6.10	0.07\\
6.20	0.07\\
6.30	0.06\\
6.40	0.06\\
6.50	0.06\\
6.60	0.06\\
6.70	0.06\\
6.80	0.06\\
6.90	0.06\\
7.00	0.06\\
7.10	0.06\\
7.20	0.06\\
7.30	0.06\\
7.40	0.06\\
7.50	0.06\\
7.60	0.05\\
7.70	0.05\\
7.80	0.05\\
7.90	0.04\\
8.00	0.04\\
8.10	0.04\\
8.20	0.03\\
8.30	0.03\\
8.40	0.03\\
8.50	0.03\\
8.60	0.02\\
8.70	0.02\\
8.80	0.02\\
8.90	0.03\\
9.00	0.02\\
9.10	0.01\\
9.20	-0.00\\
9.30	-0.01\\
9.40	-0.01\\
9.50	-0.01\\
9.60	-0.00\\
9.70	0.00\\
9.80	0.02\\
9.90	0.02\\
10.00	0.02\\
10.10	0.01\\
10.20	0.01\\
10.30	-0.00\\
10.40	-0.01\\
10.50	-0.01\\
10.60	-0.01\\
10.70	0.00\\
10.80	0.01\\
10.90	0.01\\
11.00	0.00\\
11.10	0.01\\
11.20	0.01\\
11.30	0.02\\
11.40	0.02\\
11.50	0.02\\
11.60	0.02\\
11.70	0.03\\
11.80	0.02\\
11.90	0.01\\
12.00	0.01\\
12.10	0.01\\
12.20	0.02\\
12.30	0.02\\
12.40	0.02\\
12.50	0.02\\
12.60	0.01\\
12.70	0.01\\
12.80	0.01\\
12.90	0.01\\
13.00	0.01\\
13.10	0.01\\
13.20	0.01\\
13.30	0.01\\
13.40	0.00\\
13.50	0.00\\
13.60	-0.00\\
13.70	-0.01\\
13.80	-0.02\\
13.90	-0.02\\
14.00	-0.02\\
14.10	-0.02\\
14.20	-0.02\\
14.30	-0.02\\
14.40	-0.01\\
14.50	-0.01\\
14.60	-0.01\\
14.70	-0.01\\
14.80	-0.01\\
14.90	-0.01\\
15.00	-0.01\\
15.10	-0.00\\
15.20	0.00\\
15.30	-0.00\\
15.40	-0.00\\
15.50	-0.00\\
15.60	-0.01\\
15.70	-0.01\\
15.80	-0.01\\
15.90	-0.01\\
16.00	-0.01\\
16.10	-0.01\\
16.20	-0.01\\
16.30	-0.02\\
16.40	-0.02\\
16.50	-0.02\\
16.60	-0.03\\
16.70	-0.03\\
16.80	-0.03\\
16.90	-0.03\\
17.00	-0.04\\
17.10	-0.04\\
17.20	-0.04\\
17.30	-0.04\\
17.40	-0.05\\
17.50	-0.05\\
17.60	-0.05\\
17.70	-0.05\\
17.80	-0.05\\
17.90	-0.05\\
18.00	-0.05\\
18.10	-0.05\\
18.20	-0.06\\
18.30	-0.06\\
18.40	-0.07\\
18.50	-0.07\\
18.60	-0.08\\
18.70	-0.08\\
18.80	-0.09\\
18.90	-0.10\\
19.00	-0.11\\
19.10	-0.11\\
19.20	-0.12\\
19.30	-0.12\\
19.40	-0.12\\
19.50	-0.12\\
19.60	-0.12\\
19.70	-0.12\\
19.80	-0.12\\
19.90	-0.13\\
20.00	-0.13\\
20.10	-0.13\\
20.20	-0.13\\
20.30	-0.13\\
20.40	-0.13\\
20.50	-0.13\\
20.60	-0.14\\
20.70	-0.14\\
20.80	-0.14\\
20.90	-0.15\\
21.00	-0.15\\
21.10	-0.15\\
21.20	-0.15\\
21.30	-0.15\\
21.40	-0.15\\
21.50	-0.16\\
21.60	-0.15\\
21.70	-0.15\\
21.80	-0.14\\
21.90	-0.13\\
22.00	-0.13\\
22.10	-0.13\\
22.20	-0.12\\
22.30	-0.11\\
22.40	-0.10\\
22.50	-0.10\\
22.60	-0.09\\
22.70	-0.08\\
22.80	-0.07\\
22.90	-0.06\\
23.00	-0.05\\
23.10	-0.04\\
23.20	-0.02\\
23.30	-0.01\\
23.40	-0.01\\
23.50	-0.00\\
23.60	0.01\\
23.70	0.02\\
23.80	0.02\\
23.90	0.01\\
24.00	0.01\\
24.10	0.01\\
24.20	0.01\\
24.30	0.01\\
24.40	0.01\\
24.50	0.02\\
24.60	0.02\\
24.70	0.03\\
24.80	0.03\\
24.90	0.03\\
25.00	0.04\\
25.10	0.04\\
25.20	0.04\\
25.30	0.04\\
25.40	0.04\\
25.50	0.05\\
25.60	0.05\\
25.70	0.05\\
25.80	0.05\\
25.90	0.05\\
26.00	0.06\\
26.10	0.06\\
26.20	0.06\\
26.30	0.06\\
26.40	0.07\\
26.50	0.07\\
26.60	0.07\\
26.70	0.06\\
26.80	0.06\\
26.90	0.06\\
27.00	0.05\\
27.10	0.05\\
27.20	0.05\\
27.30	0.06\\
27.40	0.06\\
27.50	0.06\\
27.60	0.07\\
27.70	0.07\\
27.80	0.08\\
27.90	0.08\\
28.00	0.09\\
28.10	0.09\\
28.20	0.10\\
28.30	0.10\\
28.40	0.10\\
28.50	0.11\\
28.60	0.11\\
28.70	0.11\\
28.80	0.11\\
28.90	0.11\\
29.00	0.11\\
29.10	0.11\\
29.20	0.11\\
29.30	0.11\\
29.40	0.10\\
29.50	0.10\\
29.60	0.09\\
29.70	0.09\\
29.80	0.09\\
29.90	0.09\\
30.00	0.09\\
30.10	0.10\\
30.20	0.10\\
30.30	0.10\\
30.40	0.10\\
30.50	0.10\\
30.60	0.09\\
30.70	0.08\\
30.80	0.08\\
30.90	0.08\\
31.00	0.08\\
31.10	0.08\\
31.20	0.08\\
31.30	0.07\\
31.40	0.07\\
31.50	0.07\\
31.60	0.07\\
31.70	0.07\\
31.80	0.07\\
31.90	0.07\\
32.00	0.07\\
32.10	0.07\\
32.20	0.07\\
32.30	0.07\\
32.40	0.07\\
32.50	0.06\\
32.60	0.07\\
32.70	0.06\\
32.80	0.06\\
32.90	0.06\\
33.00	0.06\\
33.10	0.06\\
33.20	0.05\\
33.30	0.05\\
33.40	0.04\\
33.50	0.04\\
33.60	0.04\\
33.70	0.04\\
33.80	0.03\\
33.90	0.03\\
34.00	0.03\\
34.10	0.02\\
34.20	0.02\\
34.30	0.01\\
34.40	0.01\\
34.50	-0.00\\
34.60	-0.01\\
34.70	-0.01\\
34.80	-0.02\\
34.90	-0.02\\
35.00	-0.02\\
35.10	-0.02\\
35.20	-0.02\\
35.30	-0.03\\
35.40	-0.03\\
35.50	-0.04\\
35.60	-0.04\\
35.70	-0.04\\
35.80	-0.04\\
35.90	-0.04\\
36.00	-0.04\\
36.10	-0.04\\
36.20	-0.04\\
36.30	-0.04\\
36.40	-0.04\\
36.50	-0.04\\
36.60	-0.05\\
36.70	-0.05\\
36.80	-0.05\\
36.90	-0.05\\
37.00	-0.04\\
37.10	-0.04\\
37.20	-0.04\\
37.30	-0.04\\
37.40	-0.04\\
37.50	-0.04\\
37.60	-0.04\\
37.70	-0.04\\
37.80	-0.04\\
37.90	-0.04\\
38.00	-0.04\\
38.10	-0.04\\
38.20	-0.05\\
38.30	-0.04\\
38.40	-0.04\\
38.50	-0.04\\
38.60	-0.05\\
38.70	-0.05\\
38.80	-0.06\\
38.90	-0.06\\
39.00	-0.06\\
39.10	-0.06\\
39.20	-0.07\\
39.30	-0.07\\
39.40	-0.08\\
39.50	-0.08\\
39.60	-0.09\\
39.70	-0.09\\
39.80	-0.10\\
39.90	-0.10\\
40.00	-0.11\\
40.10	-0.11\\
40.20	-0.10\\
40.30	-0.10\\
40.40	-0.11\\
40.50	-0.11\\
40.60	-0.11\\
40.70	-0.11\\
40.80	-0.11\\
40.90	-0.11\\
41.00	-0.10\\
41.10	-0.16\\
41.20	-0.33\\
41.30	-0.52\\
41.40	-0.70\\
41.50	-0.88\\
41.60	-1.08\\
41.70	-1.25\\
41.80	-1.41\\
41.90	-1.57\\
42.00	-1.74\\
42.10	-1.87\\
42.20	-2.04\\
42.30	-2.17\\
42.40	-2.29\\
42.50	-2.43\\
42.60	-2.57\\
42.70	-2.68\\
42.80	-2.78\\
42.90	-2.88\\
43.00	-2.96\\
43.10	-3.00\\
43.20	-3.04\\
43.30	-3.08\\
43.40	-3.07\\
43.50	-3.08\\
43.60	-3.05\\
43.70	-3.01\\
43.80	-2.96\\
43.90	-2.92\\
44.00	-2.86\\
44.10	-2.81\\
44.20	-2.76\\
44.30	-2.72\\
44.40	-2.70\\
44.50	-2.66\\
44.60	-2.64\\
44.70	-2.22\\
44.80	-1.58\\
44.90	-1.03\\
45.00	-0.61\\
45.10	-0.29\\
45.20	-0.08\\
45.30	0.05\\
45.40	0.13\\
45.50	0.17\\
45.60	0.21\\
45.70	0.22\\
45.80	0.24\\
45.90	0.25\\
46.00	0.25\\
46.10	0.25\\
46.20	0.26\\
46.30	0.27\\
46.40	0.28\\
46.50	0.29\\
46.60	0.31\\
46.70	0.31\\
46.80	0.33\\
46.90	0.34\\
47.00	0.33\\
47.10	0.33\\
47.20	0.33\\
47.30	0.33\\
47.40	0.33\\
47.50	0.33\\
47.60	0.33\\
47.70	0.31\\
47.80	0.30\\
47.90	0.30\\
48.00	0.29\\
48.10	0.28\\
48.20	0.28\\
48.30	0.27\\
48.40	0.27\\
48.50	0.27\\
48.60	0.26\\
48.70	0.25\\
48.80	0.24\\
48.90	0.24\\
49.00	0.23\\
49.10	0.23\\
49.20	0.22\\
49.30	0.23\\
49.40	0.22\\
49.50	0.22\\
49.60	0.22\\
49.70	0.21\\
49.80	0.21\\
49.90	0.21\\
50.00	0.20\\
50.10	0.20\\
50.20	0.20\\
50.30	0.19\\
50.40	0.19\\
50.50	0.19\\
50.60	0.19\\
50.70	0.19\\
50.80	0.19\\
50.90	0.18\\
51.00	0.18\\
51.10	0.18\\
51.20	0.17\\
51.30	0.17\\
51.40	0.16\\
51.50	0.16\\
51.60	0.16\\
51.70	0.16\\
51.80	0.16\\
51.90	0.15\\
52.00	0.15\\
52.10	0.13\\
52.20	0.13\\
52.30	0.13\\
52.40	0.14\\
52.50	0.14\\
52.60	0.14\\
52.70	0.14\\
52.80	0.13\\
52.90	0.12\\
53.00	0.12\\
53.10	0.11\\
53.20	0.10\\
53.30	0.09\\
53.40	0.08\\
53.50	0.08\\
53.60	0.07\\
53.70	0.06\\
53.80	0.06\\
53.90	0.05\\
54.00	0.05\\
54.10	0.04\\
54.20	0.04\\
54.30	0.04\\
54.40	0.04\\
54.50	0.03\\
54.60	0.03\\
54.70	0.02\\
54.80	0.02\\
54.90	0.02\\
55.00	0.02\\
55.10	0.01\\
55.20	0.01\\
55.30	0.01\\
55.40	0.01\\
55.50	0.01\\
55.60	0.01\\
55.70	0.01\\
55.80	0.01\\
55.90	0.01\\
56.00	0.00\\
56.10	-0.00\\
56.20	-0.00\\
56.30	-0.01\\
56.40	-0.00\\
56.50	-0.00\\
56.60	-0.00\\
56.70	-0.00\\
56.80	-0.01\\
56.90	-0.01\\
57.00	-0.01\\
57.10	-0.02\\
57.20	-0.02\\
57.30	-0.02\\
57.40	-0.02\\
57.50	-0.03\\
57.60	-0.03\\
57.70	-0.03\\
57.80	-0.04\\
57.90	-0.04\\
58.00	-0.05\\
58.10	-0.05\\
58.20	-0.05\\
58.30	-0.05\\
58.40	-0.06\\
58.50	-0.07\\
58.60	-0.07\\
58.70	-0.07\\
58.80	-0.08\\
58.90	-0.08\\
59.00	-0.08\\
59.10	-0.09\\
59.20	-0.09\\
59.30	-0.10\\
59.40	-0.10\\
59.50	-0.11\\
59.60	-0.11\\
59.70	-0.11\\
59.80	-0.11\\
59.90	-0.11\\
60.00	-0.12\\
60.10	-0.12\\
60.20	-0.12\\
60.30	-0.12\\
60.40	-0.13\\
60.50	-0.13\\
60.60	-0.13\\
60.70	-0.12\\
60.80	-0.11\\
60.90	-0.11\\
61.00	-0.11\\
61.10	-0.11\\
61.20	-0.11\\
61.30	-0.11\\
61.40	-0.11\\
61.50	-0.10\\
61.60	-0.10\\
61.70	-0.10\\
61.80	-0.10\\
61.90	-0.10\\
62.00	-0.10\\
62.10	-0.09\\
62.20	-0.09\\
62.30	-0.09\\
62.40	-0.09\\
62.50	-0.09\\
62.60	-0.09\\
62.70	-0.08\\
62.80	-0.08\\
62.90	-0.07\\
63.00	-0.07\\
63.10	-0.07\\
63.20	-0.07\\
63.30	-0.07\\
63.40	-0.07\\
63.50	-0.07\\
63.60	-0.06\\
63.70	-0.06\\
63.80	-0.06\\
63.90	-0.06\\
64.00	-0.06\\
64.10	-0.06\\
64.20	-0.06\\
64.30	-0.06\\
64.40	-0.06\\
64.50	-0.06\\
64.60	-0.06\\
64.70	-0.07\\
64.80	-0.07\\
64.90	-0.07\\
65.00	-0.06\\
65.10	-0.06\\
65.20	-0.06\\
65.30	-0.06\\
65.40	-0.05\\
65.50	-0.05\\
65.60	-0.04\\
65.70	-0.03\\
65.80	-0.03\\
65.90	-0.02\\
66.00	-0.01\\
66.10	-0.01\\
66.20	-0.00\\
66.30	-0.00\\
66.40	-0.00\\
66.50	-0.01\\
66.60	-0.00\\
66.70	-0.00\\
66.80	0.00\\
66.90	-0.00\\
67.00	0.00\\
67.10	0.00\\
67.20	0.00\\
67.30	0.01\\
67.40	0.01\\
67.50	-0.01\\
67.60	-0.02\\
67.70	-0.02\\
67.80	-0.02\\
67.90	-0.02\\
68.00	-0.03\\
68.10	-0.03\\
68.20	-0.03\\
68.30	-0.03\\
68.40	-0.04\\
68.50	-0.03\\
};

\end{axis}
\end{tikzpicture}%
        \caption{The heading rate command from the weighted arithmetic mean approach in the tunnel blockage environment navigation, while the \gls{nmpc} avoids obstacles.}
       \label{fig:headrate2019-09-11-13-01-57}
\end{figure}
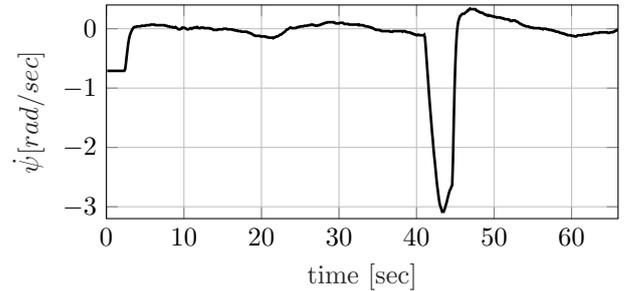

The generated heading rate command is presented in Figure~\ref{fig:headrate2019-09-11-13-01-57}, while due to strong magnetic field and disturbances in the \gls{imu}, a high peak heading rate command is instantaneously calculated around $\unit[42]{s}$ as shown in the figure, without affecting the navigation task.

%%%%%%%%%%%%%%%%%%%%%%%%%%%%%%%%%%%%%%%%%%%%%%%%
\section{Conclusions} \label{sec:Conclusions}
%%%%%%%%%%%%%%%%%%%%%%%%%%%%%%%%%%%%%%%%%%%%%%%%
This work presented a \gls{mav} navigation scheme, applied in subterranean environments, addressing the challenges of collision avoidance in such harsh environments. The proposed system is developed around a fast \gls{nmpc} framework, where the \gls{mav} is considered as a floating object following velocity references in the $x-y$ plane and position references for the altitude. The \gls{nmpc} integrates collision avoidance constraints using the kinematics of the obstacles around the \gls{mav}, using measurements from Lidar sensors. Moreover, a novel approach for correcting the heading of the \gls{mav} towards the center of the mine tunnel is proposed. Experimental trials performed in an underground mine in Sweden demonstrate the performance of the developed navigation method.

\bibliography{mybib}
\end{document}